\documentclass[journal]{IEEEtran}
\ifCLASSINFOpdf
\else
\fi
\usepackage{bbold}
\usepackage{ulem}
\usepackage{tikz}
\usepackage{microtype}
\usepackage{graphicx}
\usepackage{hyperref}
\usepackage{amsmath}
\usepackage{amssymb}
\usepackage{mathtools}
\usepackage{amsthm}
\usepackage{subcaption}
\usepackage{diagbox}
\newcommand{\NFA}{\textsc{NFA}}
\newcommand{\Sigm}{\textsc{Sigm}_{\alpha}}
\usepackage{amsfonts}
\usepackage{fancyhdr}
\usepackage{amssymb}
\usepackage{xcolor} 
\definecolor{Prune}{RGB}{99,0,60}
\usepackage{mdframed}
\usepackage{multirow} 
\usepackage{multicol} 
\usepackage{scrextend} 
\usepackage{tikz}
\usepackage{graphicx}
\usepackage[absolute]{textpos} 
\usepackage{colortbl}
\usepackage{array}
\usepackage{geometry}
\usepackage{hyperref}
\usepackage{bbold}
\usepackage{ulem}
\usepackage{amsthm}
\usepackage{stmaryrd}
\usepackage{algpseudocode} 
\usepackage{stmaryrd}
\usepackage{caption}
\usepackage{subcaption}
\usepackage{tablefootnote}
\usepackage{algorithm} 
\usepackage{upgreek}

\newcommand{\addSylv}[1]{\textcolor{black}{#1}}

\newcommand{\addCb}[1]{\textcolor{black}{#1}}

\begin{document}
%
\title{Robust infrared small target detection using self-supervised and \textit{a contrario} paradigms}
%
%
%
 \author{Alina Ciocarlan,
        Sylvie Le Hégarat-Mascle,
        Sidonie Lefebvre
        and~Arnaud Woiselle

}
\maketitle

\begin{abstract}
Detecting small targets in infrared images poses significant challenges in defense applications due to the presence of complex backgrounds and the small size of the targets. Traditional object detection methods often struggle to balance high detection rates with low false alarm rates, especially when dealing with small objects. In this paper, we introduce a novel approach that combines \textit{a contrario} paradigm with Self-Supervised Learning (SSL) to improve Infrared Small Target Detection (IRSTD). On the one hand, the integration of an \textit{a contrario} criterion into a YOLO detection head enhances feature map responses for small and \textit{unexpected} objects while effectively controlling false alarms. On the other hand, we explore SSL techniques to overcome the challenges of limited annotated data, common in IRSTD tasks. Specifically, we benchmark several representative SSL strategies for their effectiveness in improving small object detection performance. Our findings show that instance discrimination methods outperform masked image modeling strategies when applied to YOLO-based small object detection. Moreover, the combination of the \textit{a contrario} and SSL paradigms leads to significant performance improvements, narrowing the gap with state-of-the-art segmentation methods and even outperforming them in frugal settings. This \addCb{two}-pronged approach offers a robust solution for improving IRSTD performance, particularly under challenging conditions.
\end{abstract}


%
\IEEEpeerreviewmaketitle

\section{Introduction}
%
%
%
%

Infrared Small Target Detection (IRSTD) is a highly challenging but critical task in defense. The main difficulties arise from (i)~the extremely small size of the targets, often occupying less than 20 pixels, (ii)~the presence of complex and highly textured backgrounds \addSylv{that} can result in numerous false alarms, and (iii)~the challenging learning conditions. These conditions include training on small and heavily class-imbalanced datasets.


In the literature for IRSTD, several methods have been proposed to address some of the above issues. The increasing availability of infrared small target detection datasets has \addSylv{spurred} the development of deep learning-based methods \addSylv{that} have demonstrated their effectiveness in extracting non-linear features from large annotated datasets. For example, dense nested U-shaped architectures (e.g., DNANet~\cite{li2022dense}) and specific multi-scale fusion modules (e.g., LSPM~\cite{huang2021infrared}) have been introduced, which effectively limit the information loss \addSylv{on} small targets and lead to good performance on several IRSTD benchmarks. Some methods also include local and large-scale attention mechanisms in order to limit the confusion between targets and background elements (e.g., AGPCNet~\cite{zhang2023attention}). 

Note however that State-Of-The-Art (SOTA) IRSTD methods all rely on segmentation networks, and a major issue \addSylv{with} relying on such neural networks for object detection is that object fragmentation can occur when the segmentation map \addSylv{is binarized}. This issue often leads to numerous undesired false alarms. Object detection algorithms \addSylv{such as} Faster-RCNN~\cite{ren2015faster} or YOLO~\cite{redmon2016you} mitigate this risk by explicitly localizing objects through bounding box regression. 
However, traditional object detection methods struggle to balance a high detection rate with a low false alarm rate. While some existing approaches have improved feature map responses for small targets~\cite{mou2023yolo, li2023yolosr, yang2024eflnet}, they \addSylv{often} fail to manage false alarms caused by background elements, and no rigorous comparison with SOTA IRSTD methods \addSylv{has been made}. Moreover, these methods typically do not leverage the \textit{unexpected} nature of small objects relative to the background, which could be addressed using an anomaly detection approach. Such a criterion helps in distinguishing small objects as unexpected patterns against the background, effectively reducing the Number of False Alarms (NFA) induced by background noise. As demonstrated in recent studies~\cite{ciocarlan2024contrario}, this approach can achieve a better balance between precision and detection rate. Lastly, traditional object detectors face difficulties when dealing with class-imbalanced or limited datasets, which is a common and challenging condition in IRSTD.

In the following, we propose to combine both \textit{a contrario} and self-supervised learning paradigms to improve IRSTD. For this purpose, we propose a new YOLO detection head that integrates an \textit{a contrario} criterion to perform the detection of small targets. \addSylv{Inspired by} perception theory and anomaly detection methods, this criterion allows us to introduce an \textit{a priori} on small targets -- specifically, that they are \textit{unexpected} -- and to control the number of false alarms. Then, we explore the benefits of using self-supervised learning (SSL) to initialize the backbone of a modified version of YOLO. The aim of this unsupervised pre-training paradigm is to help the network learning features or invariances that are relevant \addSylv{to} the downstream task. 
One of the motivations \addSylv{stems} from the fact that SSL methods have been shown in the literature to improve SOTA performance for many use cases. More specifically, SSL allows the network to learn general features from large unlabeled datasets which, when transferred to a final task, improve\addSylv{s} performance despite difficult fine-tuning conditions (e.g., little annotated data or few computational resources). However, the use cases considered in the literature mainly concern classification or the detection of objects of medium to large size. This raises the following question: do these conclusions transfer well to the detection of small targets? 

In this study, we \addSylv{show} that SSL is indeed beneficial for IRSTD, particularly when combined with the \textit{a contrario} detection head, leading to more robust results in challenging conditions (e.g., frugal setting). Our contributions are three-fold:
\begin{itemize}
    \item \textbf{YOLO + $\NFA_{\mathcal{N}}$ detection head}: We introduce a new YOLO detection head that integrates a pixel-level \textit{a contrario} criterion, creating a network we call YOLO + $\NFA_{\mathcal{N}}$. This detection head can be integrated into any YOLO detector and is more robust to challenging conditions, such as frugal settings or complex backgrounds.
    \item \textbf{SSL pre-training strategies and IRSTD}: We analyze the impact of different SSL pre-training strategies for IRSTD and show that instance discrimination methods are more effective than Masked Image Modeling (MIM) methods when used ``as is'' \addSylv{to initialize} a convolution-based backbone. 
    \item \textbf{New robust and SOTA results for IRSTD}: Last but not least, we show that \addSylv{the combination of} both SSL and \textit{a contrario} paradigms not only narrows the performance gap with SOTA segmentation methods on very challenging IRSTD datasets, but can also outperform them \addSylv{by} a large margin, especially in a frugal setting. 
\end{itemize}

\section{Related works}

\subsection{\textit{A contrario} paradigm}
In order to explicitly take advantage of the amount of information available on the background to discriminate small targets, it is interesting to consider reasoning used in anomaly detection, such as \textit{a contrario} reasoning. \textit{A contrario} paradigm, introduced in~\cite{desolneux2007gestalt}, consists in the rejection of a naive model that describes a destruct\addSylv{u}red background. Such reasoning originates from theories of perception, in particular \addSylv{from} Gestalt theory. The threshold used to reject the background hypothesis \addSylv{allows us} to control the Number of False Alarms (NFA). The latter can be defined as the product between the total number of tested \textit{objects} and the tail distribution of the law followed by the chosen naive model. Depending on the type of object we consider, several \textit{a contrario} formulation\addSylv{s} can be \addSylv{considered}. 
For example, the most commonly used and straightforward naive model for gray-level feature maps is the Gaussian distribution of the pixel gray-level values~\cite{ipol.2019.263, vidal2019aggregated}. In this case, high intensity pixels (compared to the global statistics of the image) are more likely to belong to the target class. Such a criterion has been integrated into segmentation neural networks in~\cite{ciocarlan2024contrario} and ha\addSylv{s} led to great performance \addSylv{in} tiny object detection. When dealing with binary images, it is common to use the uniform spatial distribution of ``true'' pixels in the image grid as a naive model. \addSylv{Then, it boils down to assuming a binomial distribution for} the number of pixels falling into a parametric shape describing our object of interest~\cite{desolneux2003grouping, HegaratMascle2019}. The advantage of\textit{ a contrario} methods over other statistical tests (e.g., the family-wise error rate control) is that, in contrast to the latter, there is no control of the probability, but rather of the number of false alarms. Consequently, an increase in the size of the image has no effect on the total number of false alarms in the image. In addition, the explicit use of background information to discriminate targets alleviates the constraint of requiring a large number of target samples, and therefore a large number of data, in order to achieve high performance.

\subsection{Self-supervised learning and object detection}
SSL is a SOTA approach for performing unsupervised pre-training on large unlabeled datasets, and is a particularly active area of research. It relies on a pretext training task \addSylv{able to generate its own ground truth (e.g., pseudo-labels)}, and such \addSylv{a} strategy helps the network to learn invariances and latent patterns in the data. Several pretext tasks have been proposed in the literature, which we can divide into two main categories: instance discrimination methods, and Masked Image Modeling (MIM) methods.  

Instance discrimination methods aim at modeling the decision \addSylv{boundaries} between sub-sets of data represented in the latent space. Fundamental methods such as MoCov2~\cite{he2020momentum} or DINO~\cite{caron2021emerging} consider images as instances, and perform inter-image discrimination. Concretely, given two transformed images, if these images are augmented views (called positive samples or pairs) of the same anchor image, \addSylv{then} their features are forced to be represented similarly in the latent space. Teaching a network to identify whether two images come from the same anchor image forces the encoder to learn general and representative features of a given image while being invariant to the augmentations used to create \addSylv{the} augmented views. Common data augmentations include random rotation, color jitter or \addSylv{G}aussian noise and \addSylv{G}aussian blur. Data augmentations should be \addSylv{carefully} chosen according to the desired invariances (e.g., invariance to color, illumination, or occlusion) for the downstream task. Note however that most instance discrimination methods assume that the images are semantically consistent, which may impair the performance for dense prediction tasks. 
For this purpose, some local instance discrimination methods have been proposed in the literature by, for example, designing \addSylv{data-augmentations at the} object or region-level~\cite{xiao2021region}, or applying instance discrimination loss at a local level (e.g., per pixel~\cite{wang2021dense}). In contrast to global instance discrimination techniques, which treat an image as a single instance, local instance discrimination methods appear more suitable for object detection, as they allow for the extraction of more local features. 

\addSylv{Unlike} instance discrimination methods, the objective of MIM methods is to recover corrupted images. The underlying hypothesis is that, if a network is able to reconstruct severely masked information, then it ``understands'' the semantics in the image. One strong invariance learned by such methods is occlusion invariance. Moreover, by reconstructing the corrupted data, image modeling methods better exploit local features~\cite{xie2023revealing}. Famous MIM methods include Masked AutoEncoders (MAE, ~\cite{he2022masked}) and SimMIM~\cite{xie2022simmim}, which have shown impressive performance on several object detection benchmarks, especially when combined with Vision Transformer~\cite{dosovitskiy_image_2021} encoders.

\section{Proposed method}

\subsection{Integrating an \textit{a contrario} criterion into YOLO}

\begin{figure*}[h]
    \centering
    \includegraphics[width=16cm]{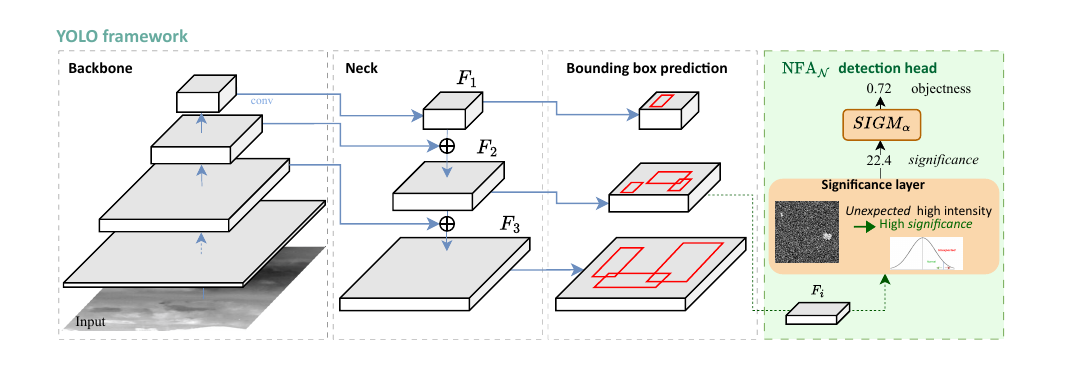}
    \caption{ Integration of our pixel-level criterion into a YOLO framework, through the $\NFA_{\mathcal{N}}$ detection head. This module can be added on top of any YOLO. }
    \label{fig:yolo_NFA_pix_arch}
\end{figure*}

In this section, we describe our method for integrating an \textit{a contrario} criterion into a YOLO detection head. Specifically, the \textit{a contrario} test is \addSylv{used} to re-estimate the objectness scores predicted by the YOLO detection head for each bounding box. 
Traditionally, 
\textit{a contrario} criteria have been used to detect \addSylv{objects based on a specific feature such as alignment}~\cite{desolneux2007gestalt}, \addSylv{that can be easily  measured, i.e., by} counting the number of points within the parametric shape \addSylv{(strip for alignment) that characterizes the object's geometry}. 
Similarly, one might consider counting the points within the bounding box predicted by YOLO. 
However, incorporating such a counting operation into the training loop of a neural network in a differentiable manner is challenging. This process typically requires thresholding to discretize the feature map, which breaks the continuity of the forward pass and can lead to an unstable training process. To address this issue, we propose to rely on a simpler pixel-level \textit{a contrario} formulation. Instead of counting points within the bounding box, we estimate the objectness score for each bounding box using \addSylv{only} its center, thereby ignoring its spatial extent and density. While this may seem limiting, we argue that this approach is actually efficient, particularly in the context of small object detection where the spatial extent of the bounding box is very small. Indeed, we will show that this strategy works \addSylv{very} well in practice.  

Concretely, \addSylv{our} naive assumption $H_0$ \addSylv{is} that the background noise \addSylv{follows a Gaussian distribution} and \addSylv{we} rely on the multi-channel NFA criterion introduced in~\cite{ciocarlan2024contrario} for segmentation networks, which is defined for a centered input $X_i$ with $K$ channels as:
\begin{equation}
    \NFA_{\mathcal{N}}(x_i) = \frac{\eta_{test}}{\Gamma(K/2)} \Gamma(\frac{K}{2},\frac{1}{2} ||\Sigma^{-1/2} x_i||^2_2),
    \label{eq:NFA_form_gamma}
\end{equation}
 where $\Gamma(.)$ and $\Gamma(.,.)$ are the Gamma and upper incomplete Gamma functions respectively, $\Sigma$ represents the covariance matrix of the centered variable $X_i$, and $\eta_{test}$ is the number of tested pixels (i.e., the total number of pixels within an image).

The integration of this $\NFA_{\mathcal{N}}$ criterion is illustrated on Figure~\ref{fig:yolo_NFA_pix_arch}. To do so, we first modify the YOLO detection head by separately predicting the bounding box coordinates, the classification scores and the objectness scores. \addSylv{Then}, after the original convolution step for predicting the objectness score, \addSylv{we add} our NFA detection head. The latter consists in computing the \textit{significance} for each box, defined as $-\log(\NFA_{\mathcal{N}})$. Finally, we apply the $\Sigm$ activation function defined in~\cite{ciocarlan2024contrario} to obtain an objectness score that ranges between $0$ and $1$. Note that since the YOLO version we consider is multi-scale, we \addSylv{get multiple} \textit{significance} maps (one for each scale). However, \addSylv{since we define a constant value for $\eta_{test}$,} each scale has the same weight in the decision. This strategy may be sub-optimal in some cases, where many large objects need to be detected. For this purpose, we introduce some weighting coefficient\addSylv{s}, which are obtained using an attention layer, namely the ECA \addSylv{(Efficient Channel Attention~\cite{eca_net})} layer. The integration of \addSylv{this}  layer is omitted in Figure~\ref{fig:yolo_NFA_pix_arch} for simplicity.

\addSylv{Finally, note that w}e train the YOLO$+\NFA_{\mathcal{N}}$ in an end-to-end manner using the Mean Squared Error loss instead of \addSylv{the} Binary Cross Entropy loss for the objectness scores as it has shown to lead to better performance in our experiments.

\subsection{Choosing appropriate SSL initialisation}

In this work, we evaluate the benefits of SSL pre-training for IRSTD, with a particular focus on its effectiveness when combined with our YOLO + $\NFA_{\mathcal{N}}$ detector. To this end, we select a range of SSL strategies, including methods from MIM, global instance discrimination, and local instance discrimination approaches. Although local methods are expected to work better for our task, we choose to test several methods representative of different SSL strategies, including global instance discrimination methods. This allows us to assess the relevance of the different SSL strategies for our use case.
It is important to note that, in the literature, pre-trained weights are available only for ResNet-50 when dealing with convolutional backbones. Therefore, we adapt the YOLOv7-tiny architecture by replacing its original backbone with a ResNet-50. The SSL methods we consider for evaluation are as follows:
\begin{itemize}
    \item \textbf{DINO}~\cite{caron2021emerging} -- This method belongs to the global instance discrimination category, and originates from BYOL~\cite{grill2020bootstrap}, a pioneering SSL method based on self-distillation. Concretely, it consists in distilling the knowledge from a teacher branch to a student one (i.e.\addSylv{,} training the student network to predict the representations learned by the teacher). To prevent the network from collapsing, the teacher\addSylv{'s} weights are not shared with the student branch (i.e.\addSylv{,} there is no backpropagation): instead, the teacher\addSylv{'}s weights are progressively updated through an exponential moving average of the student\addSylv{'s} weights. 
    DINO further improves BYOL by smoothly discretizing the representations. \addSylv{The a}uthors also show that multi-crop augmentations are essential for improving the fine-tuning performance. \addSylv{Specifically, local-to-global correspondences are learned by providing large crops to the teacher and small crops to the student that thus is trained to interpolate context from a small crop.} 
    \item \textbf{ReSim}~\cite{xiao2021region} -- This pretext task belongs to local instance discrimination category, where region-level augmentations are considered. Specifically, a sliding window extracts, in each branch \addSylv{of a siamese network}, local features within the overlapping area between the two augmented views of the anchor sample. This creates local positive pairs that represent exactly the same spatial region in the original image (we say that the patches are geometrically aligned). The similarity between the pairs of local patches is \addSylv{thus} enforced. ReSim can be built on several SSL frameworks, including SimSiam and MoCov2. In our experiments, we choose the MoCov2 framework, as it has led to better results in the original paper. In this case, negative samples can be sampled either from non-positive regions of the same image, or from crops from other images in the dataset.
    \item \textbf{SparK}~\cite{tian2022designing} -- This pretext task belongs to MIM methods, and is an adaptation of the famous MAE~\cite{he2022masked} method to convolutional encoders. \addSylv{U}nlike ViT architectures that analyze each patch independently, CNN-based encoders perform convolutions by sliding a window, and thus the receptive field of the convolution can overlap with both masked and unmasked areas. This leads to several issues such as masked pattern vanishing or the disturbance of \addSylv{the} distribution \addSylv{of} pixel values, as explained in~\cite{tian2022designing}. SparK introduces the use of partial or sparse convolution\addSylv{s, i.e. convolutions that provide output only when the kernel center covers an active (unmasked) input site}. \addSylv{The} authors of \addSylv{the} SparK~\cite{tian2022designing} paper show that MAE pre-training with a CNN-based encoder can outperform ViT-based MAE pre-training when using sparse convolution and a modern CNN-based encoder, namely ConvX-B~\cite{liu2022convnet}. 
\end{itemize}

We use weights pre-trained on the ImageNet dataset to initialize the encoder of our YOLO + $\NFA_{\mathcal{N}}$ detector. In the literature, encoders pre-trained with SSL methods are evaluated on the downstream tasks after fine-tuning the entire classification, detection or segmentation network. However, as explained in~\cite{vasconcelos2022proper}, such a fine-tuning strategy may not be suitable for dense prediction tasks\addSylv{, i.e. tasks that output an image or a region set, such as small target detection}. This can be explained by the fact that a complex, randomly initialized detection or segmentation head has to be added on top of the encoder, and the backpropagation of these random weights can ``break'' the knowledge learned during the SSL pre-training of the encoder. To further improve the transfer learning performance on IRSTD and to investigate the benefits of each SSL method without the effects of fine-tuning, we propose to freeze the backbone layers (i.e., the ResNet layers in YOLO-R50) and to fine-tune \addSylv{only} the rest of the neural network (i.e., the YOLOv7-tiny detection head), as in~\cite{vasconcelos2022proper}.
 
\section{Experiments}
\subsection{Experimental set-up}
\subsubsection{Datasets and evaluation metrics} To assess our methods, we consider two IRSTD datasets, namely SIRST~\cite{dai2021asymmetric} and IRSTD-1k~\cite{zhang2022isnet} datasets. 

SIRST is one of the first \addSylv{publicly released} real-image infrared small target datasets, and it is widely used in the literature as a reference dataset for IRSTD. This dataset contains $427$ real mono-spectral infrared images (in NIR, SWIR or MWIR domains) with resolution $256\times256$. Moreover, $90\%$ of the images contain a single target, and most targets follow the definition of a small target proposed by \addSylv{the} SPIE \addSylv{(Society of Photo-Optical Instrumentation Engineers)}, i.e. objects having a total spatial extent of less than $80$ pixels ($9\times 9$)~\cite{1279357}. 

IRSTD-1k is a recently published dataset; it is larger than SIRST ($1000$ images of resolution $512\times512$) and contains more challenging scenes. It also contains some relatively large objects. Since our work focus\addSylv{es} on developing and evaluating methods for \textit{small} target detection, we follow what is done in~\cite{ciocarlan2024contrario} and decide to remove \addSylv{the} images that contain targets having a spatial extent larger than $90$ pixels (this represents $15\%$ of the dataset). This filtered version of IRSTD-1k dataset is referred to as ``IRSTD-850''. As YOLO networks are designed to take large image \addSylv{as} inputs, we upsample all images to the size $640\times640$ using bi-cubic interpolation. Both datasets are split into training, validation and test sets using a ratio of $60:20:20$.

For the evaluation, we focus on conventional object-level metrics, namely the F1 score and the Average Precision (AP), which computes the area under the object-level Precision-Recall curve. A detected object is counted as a true positive (TP) if it has an Intersection over Union (IoU) of at least $5\%$ with the ground truth. This low-constrained condition is due to the fact that \addSylv{for small targets,} a small shift in the number of predicted pixels leads to a large deviation in the IoU, as illustrated in \addSylv{Fig.1 of}~\cite{cheng2023towards}. 

\subsubsection{Baselines}
We compare our methods to several baselines. These include SOTA segmentation baselines such as DNANet~\cite{li2022dense} and DNIM+NFA~\cite{ciocarlan2024contrario}. DNANet consists of several nested UNets and a multiscale fusion module that\addSylv{, together,} enable the segmentation of small objects variable size. DNIM+NFA consists in introducing an \textit{a contrario} criterion on top of DNIM backbone (which is the backbone of DNANet network). We also consider YOLOv7-tiny and YOLO-R50 (YOLOv7-tiny with a ResNet-50 backbone) as YOLO baselines. All YOLO-based networks are trained from scratch on Nvidia RTX6000 GPU for $600$ epochs, with Adam optimizer~\cite{kingma2014adam}, a batch size \addSylv{equal to} $16$ and a learning rate \addSylv{equal to} $0.001$. We use the same data-augmentation functions as those proposed by default in YOLOv7-tiny implementation. For DNANet and DNIM+NFA, we use the results reported in~\cite{ciocarlan2024contrario}, since our experimental setup is identical. The results presented in the \addSylv{t}ables are averaged across three different runs.

\subsection{Mixing both SSL and \textit{a contrario} paradigms improve\addSylv{s} the baselines}
\bgroup
\def\arraystretch{1.15}
\begin{table}[h] 
\small
  \centering
\begin{tabular}{rlllll} 
  \hline
     \multirow{2}{*}{\textbf{Backbone}}  & \multirow{2}{*}{\textbf{Init.}}&  \multicolumn{2}{c}{\textbf{SIRST} } & \multicolumn{2}{c}{\textbf{IRSTD-850} }\\\cline{3-6}
      
        & & \textbf{F1 }  &$\textsc{AP}$  & \textbf{F1 }  & $\textsc{AP}$    \\
     \hline
     
     \hline
    YOLO-R50 & Scratch & 97.5 & 98.1 & 82.3 & 84.3 \\
     +$\NFA_{\mathcal{N}}$ & Scratch &  $96.9$ & $98.3$ & $87.0$ & $88.4$ \\
     \hline
   \multicolumn{4}{l}{\textcolor{darkgray}{\textit{Instance discrimination methods}}}  \\

  +$\NFA_{\mathcal{N}}$ & DINO &  $97.5$ & $\textbf{98.6}$ & $\textbf{88.4}$ & $\textbf{90.4}$ \\
     
  +$\NFA_{\mathcal{N}}$ & ReSim &  $\textbf{99.1}$ & $\textbf{98.6}$ & $\underline{87.6}$ & $\underline{90.0}$ \\
  
      \hline
    \multicolumn{4}{l}{\textcolor{darkgray}{\textit{MIM methods}}}  \\

  +$\NFA_{\mathcal{N}}$ & SparK &  $\underline{97.8}$ & $\underline{98.5}$ & $86.9$ & $89.1$ \\
    
   \hline
  \end{tabular}
  \caption{Results obtained on SIRST and IRSTD-850 datasets using a YOLO-R50 or YOLO-R50+$\NFA_{\mathcal{N}}$ with different backbone initializations (Scratch, DINO, ReSim or SparK). \addSylv{The b}est results are in bold, and \addSylv{the} second best results are underlined. }
  \label{tab:sirst_yolor50_NFA}
\end{table}
\egroup

In this section, we evaluate the benefits of combining our $\NFA_{\mathcal{N}}$ detection head with the different SSL initialization, namely DINO, ReSim and SparK. The results obtained on SIRST and IRSTD-850 datasets are presented in Table~\ref{tab:sirst_yolor50_NFA}\addSylv{. T}hey have been averaged over three distinct training sessions.
First, we can notice that integrating an \textit{a contrario} criterion into the training loop of YOLO-R50 leads to more robust and precise results as shown by the AP metric. Moreover, YOLO-R50+$\NFA_{\mathcal{N}}$ \addSylv{significantly} improves the F1 score on the challenging IRSTD-850 dataset ($+4.7\%$). 
Second, combining SSL with YOLO-R50+$\NFA_{\mathcal{N}}$ further improve\addSylv{s} the results. More specifically, YOLO-R50+$\NFA_{\mathcal{N}}$ initialized with ReSim weights achieves more than $99\%$ of F1 score, outperforming \addSylv{all} other method\addSylv{s} \addSylv{by} a \addSylv{wide} margin. It is worth noting that instance discrimination methods, whether local or global, appear to be more effective for IRSTD \addSylv{than} MIM methods. This might seem counter-intuitive, as one would expect that extracting local features would be more beneficial for detecting small objects. However, this can be explained by the fact that higher-level feature extraction complements the locality already introduced by \addSylv{the} convolutional layers. Additionally, the performance of MIM methods, which are more sensitive to image statistics due to their strong bias towards local details~\cite{park2022self}, may be limited by the domain gap between RGB \addSylv{training domain for the backbone layers, then frozen} and IR \addSylv{(application domain)}.

\begin{figure}[]
         \centering
         \includegraphics[width=9cm]{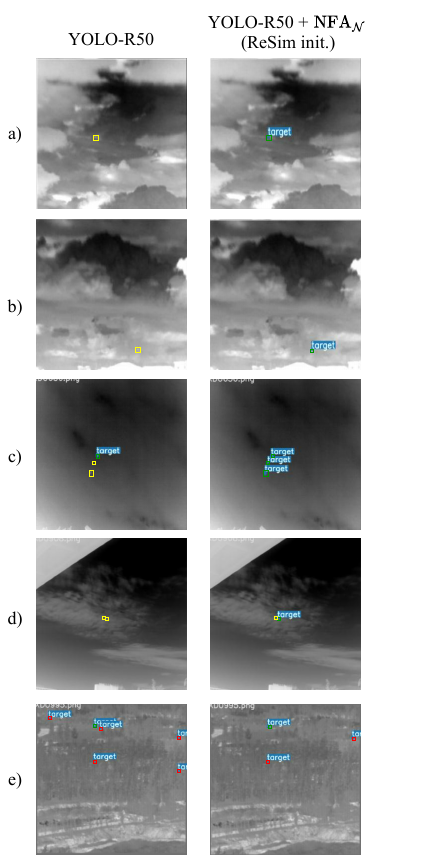}
         \caption{Qualitative results obtained with YOLO-R50 and YOLO-R50+$\NFA_{\mathcal{N}}$ initialized with ReSim weights on challenging scenes from SIRST and IRSTD-850 datasets. True positives, false positives and missed detections are circled in green, red and yellow lines, respectively.}
         \label{fig:ssl_nfa_res_vis}
     \end{figure}

Figure~\ref{fig:ssl_nfa_res_vis} provides some examples of predictions on challenging scenes from the SIRST and IRSTD-850 datasets. Our YOLO-R50+$\NFA_{\mathcal{N}}$ model initialized with ReSim weights leads to fewer missed detections compared to the baseline, particularly for very small targets, as seen in images a), b), and d). Additionally, the model shows greater robustness against false alarms, as illustrated in image e), where it effectively handles noisy backgrounds.

\subsection{What about frugal setting?}
\label{ssl_frugal_res}



    








\addSylv{For this section, we focus on the most promising method according to Table~\ref{tab:sirst_yolor50_NFA}, namely YOLO-R50+$\NFA_{\mathcal{N}}$ with ReSim initialization.} \addSylv{We aim to} evaluate \addSylv{it under more} challenging conditions, namely $25$-shot training on SIRST. Specifically, we compare YOLO-R50+$\NFA_{\mathcal{N}}$ (using either randomly initialized weights or ReSim initialization) with several baselines: 1) SOTA segmentation networks for IRSTD and 2) YOLO baselines (some with ReSim initialization). The results\addSylv{, averaged over three distinct training sessions,} are presented on Table~\ref{res_SIRST_25shot3}. 

\addSylv{The first thing to note is that while the} YOLO baselines perform particularly poorly in a frugal setting compared to SOTA segmentation networks, the integration of our $\NFA_{\mathcal{N}}$ detection head, as well as its combination with SSL pre-training, leads to very impressive results. Indeed,  YOLO-R50+$\NFA_{\mathcal{N}}$ initialized with ReSim weights \addSylv{achieves} a  F1 score of $95.4\%$ with only $10\%$ of the SIRST dataset, outperforming DNIM+NFA \addSylv{by} a wide margin ($+4.5\%$). Notably, such a strategy allows us to achieve a F1 score that is \addSylv{almost} as high as that obtained with the full SIRST dataset. It is worth noting that the NFA detection head contributes the most to \addSylv{the} good performance in frugal setting. Indeed, YOLO-R50+$\NFA_{\mathcal{N}}$ \addSylv{achieves} a  F1 score of $91.9\%$, while YOLO-R50 with the best backbone initialization leads to a F1 score of \addSylv{only} $43.6\%$. This highlights the robustness of our method in difficult training scenarios.

\bgroup
\def\arraystretch{1.15}
\begin{table}[t] 
\small
\centering
 
  \begin{tabular}{lll} 
  \hline
    \multirow{2}{*}{\textbf{Method}} & \multicolumn{2}{l}{\textbf{25-shots}}\\\cline{2-3}
    
   & \textbf{F1 }  & \textbf{AP }  \\
   \hline

   \hline
    \multicolumn{3}{l}{\textcolor{darkgray}{\textit{SOTA segmentation baselines for IRSTD}}}  \\
DNANet~\cite{li2022dense} & 73.1 & 63.7 \\
DNIM+$\NFA$~\cite{ciocarlan2024contrario} & 90.9 & 93.1 \\
\hline
    \multicolumn{3}{l}{\textcolor{darkgray}{\textit{YOLO baselines}}}  \\
YOLOv7-tiny  & 21.8 & 15.0 \\
YOLO-R50 & 26.1 & 23.1\\
 YOLO-R50 + ReSim &  43.6 & 43.7\\
\hline
    \multicolumn{3}{l}{\textcolor{darkgray}{\textit{Our methods}}}  \\
   
    YOLO-R50 + $\NFA_{\mathcal{N}}$ & \underline{91.9} & \underline{94.8}\\
 YOLO-R50 + $\NFA_{\mathcal{N}}$ + ReSim & \textbf{95.4} & \textbf{96.6}\\
   \hline
  \end{tabular}
  \caption{Results achieved in a 25-shot setting on SIRST dataset. The best results are in bold, and the second best results are underlined. }
  \label{res_SIRST_25shot3}
\end{table}
\egroup

\subsection{Comparison with SOTA segmentation networks}
\bgroup
\def\arraystretch{1.15}
\begin{table}[t] 
\small
  \centering
\begin{tabular}{lllll} 
  \hline
     \multirow{2}{*}{\textbf{Backbone init.}}  &  \multicolumn{2}{c}{\textbf{SIRST} } & \multicolumn{2}{c}{\textbf{IRSTD-850} }\\\cline{2-5}
      
        & \textbf{F1 }  &$\textbf{AP}$  & \textbf{F1 }  & $\textbf{AP}$    \\
     \hline
     
     \hline
   
   \multicolumn{5}{l}{\textcolor{darkgray}{\textit{SOTA segmentation baselines for IRSTD}}}  \\


    
  DNANet~\cite{li2022dense} &  97.1 & \underline{98.4} & \textbf{91.4} & \underline{92.4} \\
   DNIM+$\NFA$~\cite{ciocarlan2024contrario} &  $\underline{97.6}$ &  $\underline{98.4}$ & $\underline{91.3}$ & $\textbf{94.2}$ \\
 \hline
    \multicolumn{5}{l}{\textcolor{darkgray}{\textit{YOLO baselines}}}  \\
  YOLOv7-tiny & 96.5 & 97.8 &  82.2 & 85.0 \\
  YOLO-R50 & 97.5 & 98.1 & 82.3 & 84.3 \\
      \hline
    \multicolumn{5}{l}{\textcolor{darkgray}{\textit{Our method}}}  \\

   \multicolumn{1}{r}{+ $\NFA_{\mathcal{N}}$ + ReSim} &  $\textbf{99.1}$ & $\textbf{98.6}$ & $87.6$ & $90.0$ \\
   \hline
  \end{tabular}
  \caption{Overview of the performance obtained by SOTA IRSTD methods, YOLO baselines as well as our methods on SIRST and IRSTD-850 datasets. The best performance is given in bold, and the second best results are underlined.}
  \label{tab:concl_perf}
\end{table}
\egroup
In the previous sections, we have seen that both \textit{a contrario} and self-supervised paradigms have led to impressive results for IRSTD. Specifically, combining both paradigms greatly improves the YOLO baselines, especially in a frugal context, where the SOTA segmentation methods are outperformed \addSylv{by} a \addSylv{wide} margin. 
In this section, we compare our best \addSylv{candidate} method, namely YOLO-R50+$\NFA_{\mathcal{N}}$ initialized with ReSim weights, to SOTA segmentation baselines in a data-sufficient context on SIRST a\addSylv{n}d IRSTD-85 datasets. The results, shown in Table~\ref{tab:concl_perf}, indicate that our approach sets new SOTA results on the SIRST dataset: YOLO-R50 + $\NFA_{\mathcal{N}}$ initialized with ReSim weights outperforms DNANet and DNIM+NFA \addSylv{by} a \addSylv{wide} margin, achieving over $99\%$ F1 score. This is particularly noteworthy because detection networks typically struggle with small objects compared to segmentation networks.

The results on the IRSTD-850 dataset, however, are less encouraging, and incorporating this criterion into any YOLO backbone does not yield competitive performance. This is because both \addSylv{considered} YOLO baselines perform poorly on IRSTD-850 dataset. While adding the $\NFA_{\mathcal{N}}$ detection head to a YOLO backbone significantly improves its baseline performance, it still does not close the performance gap between YOLO-based networks and SOTA segmentation baselines. This can be explained by the fact that, unlike SIRST dataset, IRSTD-850 dataset presents more complex and textured backgrounds, making it particularly \addSylv{difficult} to distinguish the targets from the noisy background. Therefore, there is still room for improvement in object detectors for tiny object detection in complex environments. For instance, improvements could be made to the YOLO backbone to reduce information loss on small objects, or the SSL pre-training could be \addSylv{performed} on an in-domain dataset for better adaptation.

\section{Conclusion}
In this paper, we introduce an original approach for IRSTD that combines self-supervised learning with the \textit{a contrario} paradigm. Integrated within a YOLO detection framework, our approach significantly reduces the performance gap between traditional object detectors and SOTA segmentation methods for IRSTD. Furthermore, our YOLO + $\NFA_{\mathcal{N}}$ initialized with ReSim weights achieves new state-of-the-art results on the SIRST dataset, as well as impressive performance in a few-shot setting, demonstrating its effectiveness and potential for improving small target detection in challenging IR data. To further enhance the detection performance, it is recommended to consider adapting a Vision Transformer for small target detection. This approach has demonstrated effectiveness in modeling complex scenes, particularly when combined with SSL pre-training. Ultimately, the design of a more suitable transfer learning strategy should be explored.


\ifCLASSOPTIONcaptionsoff
  \newpage
\fi

\bibliographystyle{ieeetr}
\bibliography{bibtex/bib/IEEEabrv}
%

\end{document}